\documentclass[10pt, twocolumn,journal]{IEEEtran} 

\IEEEoverridecommandlockouts                              
\usepackage[letterpaper,top=0.75in,bottom=1in,left=0.625in,right=0.625in]{geometry}
\usepackage{amssymb}
\usepackage{bbm}
\usepackage{algorithm}
\usepackage{graphicx}
\usepackage{algorithmic}
\usepackage{booktabs}

\usepackage[colorlinks=true]{hyperref}
\hypersetup{
    bookmarksnumbered=true, bookmarksopen=true,
	unicode=true, colorlinks=true, linktoc=all, 
	linkcolor=blue, citecolor=blue, filecolor=blue, urlcolor=blue,
	pdfstartview=FitH
}

\usepackage{graphics,color} 
\usepackage{pgfplots}
\usetikzlibrary{patterns}
\usepackage{subfigure}
\usepackage{rotating}
\usepackage{graphicx}
\usepackage{siunitx}
\usepackage{cite}
\usepackage{times}
\usepackage{comment}
\usepackage{amssymb,url}
\usepackage{epsfig}
\usepackage[cmex10]{amsmath}
\usepackage{amsthm}
\usepackage{bm} 
\usepackage{breqn}
\usepackage{textcomp}
\usepackage{multicol}
\usepackage{lipsum}
\usepackage{mathtools} 
\usepackage{amsmath}
\usepackage{amsthm}
\usepackage{amsfonts}
\usepackage{amssymb}
\usepackage{breqn}

\begin{document}

\title{FedSV: Byzantine-Robust Federated Learning via Shapley Value
\thanks{This work was supported by ANR DELIGHT ANR-22-CE23-0024 }
}
\author{Khaoula Otmani\IEEEauthorrefmark{1}, Rachid El-Azouzi\IEEEauthorrefmark{1}\IEEEauthorrefmark{2}, Vincent Labatut\IEEEauthorrefmark{1} \\
\IEEEauthorrefmark{1}CERI/LIA, Université d'Avignon, Avignon, France \\
\IEEEauthorrefmark{2}Carnegie Mellon University, Pittsburgh, PA, USA \\
}
\maketitle
\thispagestyle{empty}
\pagestyle{empty}

\begin{abstract}
 In Federated Learning (FL), several clients jointly learn a machine learning model: each client maintains a local model for its local learning dataset, while a master server maintains a global model by aggregating the local models of the client devices. However, the repetitive communication between server and clients leaves room for attacks aimed at compromising the integrity of the global model, causing errors in its targeted predictions. In response to such threats on FL, various defense measures have been proposed in the literature~\cite{zhang2023survey}. In this paper, we present a powerful defense against malicious clients in FL, called FedSV, using the Shapley Value (SV), which has been proposed recently to measure user contribution in FL by computing the marginal increase of average accuracy of the model due to the addition of local data of a user. Our approach makes the identification of malicious clients more robust, since during the learning phase, it estimates the contribution of each client according to the different groups to which the target client belongs. FedSV's effectiveness is demonstrated by extensive experiments on MNIST datasets in a cross-silo context under various attacks. 
\end{abstract}

\section{Introduction}
Federated learning (FL), proposed by Google AI in 2017~\cite{McMahan2016CommunicationEfficientLO}, is a distributed machine learning paradigm that enables multiple clients to collaboratively train a model under the orchestration of a central server, while preserving the confidentiality of their local data. To preserve privacy, clients share their local models rather than their local data with the central server. FL systems are expected to grow exponentially, with each system itself containing a large number of small devices in different geographical regions. Moreover, powerful GPUs have become increasingly accessible, allowing the possibility to deploy larger models, which accelerate the deployment of FL. This growing demand for FL technology opens new challenges, in addition to those that appear in traditional machine learning~\cite{He2015DeepRL}. FL has been widely applied to a variety of real-world applications, including keyword spotting~\cite{Leroy2019Federated}, prediction of activity on mobile devices~\cite{Xu2021Federated} and disease identification in healthcare~\cite{Li2019RSA}. Despite FL's collaborative learning capability, it generally involves the distribution of heterogeneous (non-IID) data between clients, and distributed learning naturally leads to repetitive synchronization between server and clients. However, the global model can be manipulated by malicious clients even if only one client device is compromised~\cite{Dong2018Byzantine}. This is why research has intensified recently on federated learning, by proposing aggregating rules (e.g. truncated mean~\cite{Dong2018Byzantine}, median~\cite{Pillutla2022Robust} and bucketing technique~\cite{Zhu2023Byzantine}), which aim to resist the Byzantine failures of certain devices.

Attacks in FL are poisoning attacks~\cite{Baruch2019Circumventing}, where malicious nodes seek to poison the globally trained model by injecting the poisoned instances in the training data~\cite{Jagielski2018Manipulating,  Baruch2019Circumventing}, or, backdoor attacks, where malicious nodes inject a backdoor into the learning phase of the local model to induce misclassification  towards some targeted classes~\cite{Duc2021Flame}. Defensive measures developed in the literature against such attacks can be classified into two categories: robust aggregation~\cite{Zhu2023Byzantine} and anomaly detection~\cite{xhemrishi2023fedgt}. Robust aggregation techniques consist of aggregating the local models in a way that mitigates the effect of the attack, whereas anomaly detection aims at eliminating the malicious clients or corrupted local data. Although defenses based on robust aggregation are relatively efficient against malicious nodes, they are only robust against a few malicious clients~\cite{Cao2022FLCert} or require a clean, representative validation dataset on the server. With the increase in the number of malicious clients, a more robust solution is needed to detect malicious clients, making FL more robust in extreme cases~\cite{Gong2023Backdoor}. Moreover, these existing defensive approaches are unable to eliminate the negative effect on overall model accuracy of non-targeted attacks that do not alter the magnitude of model weights, such as the sign inversion attack, especially when the data is extremely heterogeneous (non-IID).  A mechanism of detection of  malicious nodes  based on the coalition game and the Shapley value has been proposed in \cite{BatFL}, but no study has been carried out to address the problem of calculating the Shapley value and how each node's Shapley value has been used to detect malicious nodes.

In this paper, we present a new mechanism for identifying malicious nodes in {\it Fed}erated learning  with {\it S}hapeley {\it V}alue (FedSV). First, we design an efficient algorithm that estimates the SV of each client and evaluates data from each client device to identify malicious nodes. The SV allows evaluating the discriminating power of each client against all other combinations of clients, which makes it a promising metric to assess whether a node is malicious or not. This helps to avoid misidentifying malicious nodes, especially when the local datasets of clients  are extremely heterogeneous (non-IID). In this way, the SV evaluation can assess the accuracy performance of different groups or coalitions and perform an appropriate test to identify the presence of malicious nodes~\cite{xhemrishi2023fedgt}. Hence, computing the SV needs to exhaustively evaluate the model performance on every subset of data nodes, which incurs prohibitive communication costs and high complexity. The direct application of the SV is unfeasible in practice due to the distributed local data in FL, and conceptually flawed due to the sequential participation of clients. Finding the exact SV for decentralized FL is, in addition, challenging when the number of nodes is high~\cite{FATIMA20081673}. To this end, we use stratified sampling and the Monte Carlo method to calculate the SV iteratively during the training period~\cite{mitchell2022sampling}. Based on the SV of clients, we develop a new strategy to select healthy clients based on the clustering strategy without prior knowledge of the number of malicious nodes in the system. This strategy shows an interesting ability to select healthy clients even when local clients datasets are extremely heterogeneous.  The main contributions of this paper are:

\begin{itemize}
    \item We propose a new efficient FL scheme, FedSV, which offers a robust defense against malicious nodes.
    \item We incorporate into FL an efficient algorithm to iteratively estimate the SV of each client device in parallel with the learning period, enabling FedSV to quickly detect malicious client devices.
    \item We propose a strategy for selecting honest clients based on the clustering technique, which involves grouping clients into different clusters based on their shapley values. 
    \item Extensive experience with real-life data demonstrates the effectiveness of FedSV compared with the existent defenses in the literature.
\end{itemize}

\section{Background and Problem Formulation}
In this section, we begin by reviewing the standard FL setup (FedAvg)~\cite{McMahan2016CommunicationEfficientLO}. We then describe the most poisonous attacks on FL, followed by an overview of Shapley Value~\cite{shapley1951notes}, a solution concept in cooperative game theory.
\begin{table}[htbp]\caption{Major notations used in the paper}
    \center
    \begin{tabular}{r c p{6cm} }
        \toprule
        $\mathcal{S}$ & $\triangleq$ & Set of client devices \\
        $N$           & $\triangleq$ & Number of clients $\mathcal{S}$ \\
        $w_k^t$       & $\triangleq$ & Weights of client device $k$ at round $t$ \\
        ${\bf w}^t$   & $\triangleq$ & Weights of the global model at round $t$ \\
        $T$             & $\triangleq$ & Number of rounds until convergence \\
        $\eta$        & $\triangleq$ & Learning rate \\
        $n_k$         & $\triangleq$ & Number of samples on client $k$ \\
        $E$           & $\triangleq$ & Number of epochs in a round \\
        \bottomrule
    \end{tabular}
    \label{tab:TableOfNotation}
\end{table}

\subsection{Federated Learning Setup}
 FL system involves several clients (also called nodes) during the training of a global model. Each client has its own data that he does not share with the other clients. Specifically, a central server maintains a \textit{global} model for all clients and each client maintains a \textit{local} model. The training steps of an FL model are as follows: (i) the clients receive the global model ${\bf w}^t$ from the server; (ii) Each client $k\in S$ trains its own local model using its local training dataset, and sends its model update $w^{t}_k$ to the server; (iii) The server aggregates the models uploaded by each client for the next round. The three steps are repeated until the global model converges.


\subsection{Common Attacks in FL}
FL systems can be the target of a wide range of attacks targeting the privacy, security or robustness of the systems~\cite{zhang2023survey}. Security and privacy both refer to the protection of the data. Robustness refers to a system remaining functional under extreme conditions. Among the attacks targeting the robustness of the model training, there are Byzantine and Backdoor attacks~\cite{zhang2023survey}. In this paper, we consider the scenario in which a fraction of FL clients are malicious or manipulated by a malicious adversary. Malicious clients can be injected into the system by adding devices controlled by the adversary, compromising a fraction of the healthy clients in order to poison the global model for a certain round and in particular at the beginning of learning. In our attack scenario, we assume that malicious clients can collude and carry out coordinated attacks against the global model. We focus on
Backdoor attacks, which are the most realistic type of attack for FL~\cite{shejwalkar2021drawing} and we assume that the malicious clients or attackers know the aggregation rule in order to increase transparency and trust of the federated learning system. We also consider the extreme case where the attacker has full knowledge of the local training datasets and local models of some FL clients. Even though this case has limited applicability in practice for FL, it still presents an extreme case of attack that is difficult to combat, especially as the fraction of malicious  nodes is high.

\subsection{Shapley Value} 
In cooperative game theory, the SV is used to distribute the value function fairly between clients. It is named after Lloyd Shapley, who introduced the concept in 1953~\cite{shapley1951notes}. It is mainly used to indicate the contribution of each client in a given coalition. Formally, a cooperative game is defined by a pair $({\mathcal{S}'}, \nu)$ where ${\mathcal{S}'}$ is a coalition containing a subset of clients  and $\nu: s^{\mid {\mathcal{S}'} \mid} \rightarrow \mathbb{R}$ is the value function of ${\mathcal{S}'}$. 
The advantage of this metric is that it takes into account the collaboration effects between the clients. 

The SV of client $i$ with respect to the value function $\nu$ is defined as the average marginal contribution of $i$ to coalition $\mathcal{S}'$ over all $\mathcal{S}' \subseteq {\mathcal{S}} \backslash \{i\}$: 
\begin{equation}
    \label{eq:sv}
    SV_i(\nu) = \sum_{\mathcal{S}' \subseteq {\mathcal{S}} \backslash \{i\}} \frac{ \mid \mathcal{S}' \mid! (N-\mid \mathcal{S}' \mid-1)!}{N!}
  [\nu(\mathcal{S}'\cup\{i\})-\nu(\mathcal{S}')],
\end{equation}
where $\mathcal{S}$ is the overall coalition containing all $N$ clients. 
The equivalent formula of SV can be rewritten as follows~\cite{mitchell2022sampling} 
\begin{eqnarray}
    \label{eq:sv1}
    SV_i(\nu) &=& \frac{1}{|N|!}\sum_{\pi \in\mathfrak{S}_d}[v([\pi]_{i-1}\cup\{i\})-v([\pi]_{i-1})]\\
    &\stackrel{\text{def}}{=}& \frac{1}{|N|!}\sum_{\pi \in\mathfrak{S}_d} f_i(\pi), 
\end{eqnarray}
where $d$ is the number of elements in a permutation of clients, $\pi$ is one permutation among all possible permutations of clients of size $d$, $[\pi]_{i-1}$ is the set of players ranked lower than $i$ in the ordering $\pi$, and $\mathfrak{S}_d$ is the set of all permutations of participants of size $d$. This formulation of SV in (\ref{eq:sv1}) describes the case where all clients join a coalition in a random order, and each client $i$ who has joined the coalition receives the marginal contribution that his participation would bring to those already in the coalition. 

Calculating the exact SV according to (\ref{eq:sv}) or (\ref{eq:sv1}) is an expensive operation, it requires the calculation of all the permutations of the clients in the system, which is done in $O(N!)$. In the next section, we propose a variant of the SV scheme to reduce SV complexity in the context of FL while building on several techniques that have been developed in the literature to reduce SV computational complexity but are more relevant to federated learning.


%
%

\section{System Overview}
In this section, we first present the FedSV meta-protocol with SV, that we will use throughout this paper. Next, we introduce our new \textsc{ EstimateSV} function, which calculates the SV of each client with low complexity. We then show how the use of SV improves the success of malicious client detection.

\subsection{Description of the main FedSV Protocol}
Our proposed FedSV protocol is quite straightforward: in each round, each client computes the gradients of its local loss functions, and then updates its local model. Before executing the FedSV, the server assigns to all clients an initial SV noted $\overline {sv}^0$ (line 3). The FedSV scheme executes the following steps repeatedly at each round $t$: (i) The server sends the global model ${\bf w}^{t}$ to all clients (line 3); (ii) Each client trains a local model during $E$ epochs using its local training dataset and sends its model update to the server (lines 4-5); (iii) The server, after receiving updates from all clients, computes the SV of all clients, $(sv_i^{t})_{i\in \mathcal{S}}$, according to our lightweight SV estimator $\textsc{EstimateSV}$ (line 8), calculates the average SV of all client $(\overline {sv}_i^{t})_{i\in \mathcal{S}}$ (line 9) , and selects a subset of clients, denoted $\mathcal{S}^t$, according to the strategy $\textsc{ClusFed}(\overline {sv}^{t})$, that takes as input the SV of clients in $\mathcal{S}$ (line 11). The global model at round $t$ is then calculated using the clients in $ \mathcal{S}^t$ (line 12). The resulting FedSV scheme is presented formally in Algorithm~\ref{FedSV}. The functions $\textsc{EstimateSV}$ and $\textsc{ClusFed}$ are respectively described in section~\ref{EstiSV} and ~\ref{CluFed}.

\begin{algorithm}[ht!]
    \caption{FedSV  Learning \label{FedSV}.}
    \small
    \begin{algorithmic}[1]
        \STATE \textbf{INPUTS:} parameters $\mathcal{S}$, $N$, $T$, $n_k$, $\eta$, $E$, $ \overline {sv}^0$, ${\bf w}^{0}$;
        \FOR{ $t=1,..,T-1$}
            \STATE Server sends ${\bf w}^{t-1}$ to all clients;
            \STATE Each client $k\in \mathcal{S}$ updates local parameter via $w_k^{t} \leftarrow w_k^{t} -\eta \Delta F_k(w_k^{t-1})$;
            \STATE  if $t \mod E =0 $ then send local model update $w_k^{t} $ to the server. 
            \STATE Server computes the SV of all clients:
            \FOR{ $i=1,..,N$}
                \STATE  $sv_i^t \longleftarrow \textsc{ EstimateSV}((w)_{k\in \mathcal{S}})$
                \STATE $\overline {sv}_i^{t} = \alpha  \overline {sv}_i^{t-1} + \beta sv_i^t$
            \ENDFOR
            \STATE Select clients using function $ \textsc{ClusFed}( \overline {sv}_i^{t} )$ 
            $\mathcal{S}^t \rightarrow \textsc{ClusFed}(\overline {sv}_i^{t} )$
            \STATE Server calculates the global model:
            ${\bf w}^{t} = \sum_{k\in  \mathcal{S}^t} \frac{n_k}{\sum_{k\in \mathcal{S}^t} n_k} w_k^{t}$
            \STATE Server sends ${\bf w}^{T}$to all clients;
        \ENDFOR
        \STATE \textbf{Return} ${\bf w}^{T}$ and set of malicious clients $S\backslash \mathcal{S}^T$.
    \end{algorithmic}
\end{algorithm}

\subsection{\textsc{EstimateSV} function}
\label{EstiSV}
The crucial importance of any approximation of SV is the quality of the solution based on the sampling method as well as the number of permutations $m$ among $N!$ possible permutations for the SV. For that, we use the probability of confidence defined in~\cite{soton2015Addressing} as follows: We say that $\widehat{SV}_i(\nu)$ is an $(\epsilon, \delta)$-approximation to the exact Shapley value $SV_i(\nu)$ of a client $i$ if 
\begin{equation}
    Pr(\mid \widehat{SV}_i(\nu) -SV_i(\nu)\mid \geq \epsilon )\geq 1- \delta.
    \label{bound}
\end{equation}
From Chebychev's inequality, the following inequality holds about $\widehat{SV}_k(\nu)$
\begin{equation}
    Pr(\mid \widehat{SV}_i(\nu) -SV_i(\nu)\mid \leq \epsilon ) \geq 1- \frac{Var[\widehat{SV}_i(\nu)]}{\epsilon^2}.
    \label{Chebychev}
\end{equation}
The smaller the variance, the more concentrated $\widehat{SV}_i(\nu)$ around $SV_i(\nu)$. In the classical Monte Carlo estimator $\widehat{SV}_i(\nu) =\frac{1}{m} \sum_{l=1}^m f_i(\pi_l)$, where the orders $\pi_l$, $l=1\cdots m$ are sampled independently with uniform distribution. Hence, the variance of $\widehat{SV}_k(\nu) $ is given by
\begin{equation}
    Var[\widehat{SV}_i(\nu) ] =\frac{ Var[f_i(\pi)]}{m}.
\end{equation}
Then, the number of samples $m$ required to satisfies the condition (\ref{bound}) is $m\geq \frac{Var[f_k(\pi)]}{\delta \epsilon^2}$. To estimate the SV of clients through \textsc{ EstimateSV} function, we use the Truncated Antithetic Monte Carlo (TAMC) method~\cite{mitchell2022sampling}. TAMC is a variance reduction technique for Monte Carlo integration where clients are taken as correlated pairs instead of standard IID. This technique is relevant to achieve substantial variance reduction, in particular when the local data of clients are non-IID. Instead, to sample the order $\pi$ independently, we use a joint distribution that preserves the uniform distribution but generates a pair of permutations $\pi$ and $\pi'$. If they are negatively correlated, the variance of $\widehat{SV}_i(\nu)$ is reduced. Indeed, let $Y = \frac{f_i(\pi)+f_i(\pi')}{2}$, thus the variance of $Y$ is $Var(Y) =\frac{1}{2} (Var[f_i(\pi)] + Cov(f_i(\pi), f_i(\pi')))$. Then, if $Cov(\pi, \pi')<0$, the variance of $Y$ is reduced. This strategy works perfectly in FL, since the negative correlation can be obtained by generating an order $\pi$ and reversing the order to obtain another order $\pi'$. This gives us two permutations that are negatively correlated. Another way of reducing the complexity of TAMC is to reduce the size of the order $\pi$, especially if we observe that the value function of a coalition changes slightly when a client is added. This assumption is observed when client data is moderately heterogeneous. But at least, this reduction remains highly effective when the data is extremely heterogeneous. 

Stratified Random Sampling (SRS) is another powerful method, which involves the division of a population into smaller groups known as \textit{strata}. In our context, the strata considered for a client $i$ are $\mathfrak{S}_i^l$, $l=1...d$, where $\mathfrak{S}_i^l$ is the set of $f_i(\pi)$ in which client $i$ is in position $l$ under permutation $\pi$. The size of each stratum is $(N-1)!$. Then, if we draw $m_l$ samples from each stratum such that $m=\sum_{l=1}^d m_l$, the variance of stratified random sampling is 
\begin{equation}
    Var_{srs} [\widehat{SV}_i(\nu)]= \frac{1}{m^2} \sum_{l=1}^d m_l \sigma_{il}^2,
    \label{var_srs}
\end{equation}
where $\sigma_{il}^2$ is the variance of stratum $\mathfrak{S}_i^l$. Note that the variance of a stratum is much lower compared to $Var_{srs} [\widehat{SV}_i(\nu)]$, which considerably reduces the variance. Now, if $r_{il}$ is the range of $f_i(\pi)$ in $\mathfrak{S}_i^l$, we have $ \sigma_{il}^2 \leq \frac{r_{il}^2}{4}$. Combining the above with (\ref{var_srs}), we obtain
\begin{equation}
    Var_{srs} [\widehat{SV}_i(\nu)] \leq \frac{ d. r_{max}^2} {4 m},
\end{equation}
where $r_{max}$ is the maximum range of all strata. Thus, the number of samples $m$ required to satisfies the condition (\ref{bound}) is $m\geq \frac{d.r_{max}^2}{4\delta \epsilon^2}$.

\subsection{$\textsc{ClusFed}$: Client Selection Strategy}
\label{CluFed}
In this section, we study the strategy used by the server to select the clients whose models will be used to calculate the global model. There are several ways to design this strategy by using the SV of each client. One of the common methods is to select clients based on their scores or utility is the Gibbs or Boltzmann distribution $\frac{exp(-u_i/\tau)}{\sum_{k\in \mathcal{S}} exp(u_k)/\tau}$, where $\tau$ is a positive parameter called the temperature. Increasing the temperature $\tau$ causes a decrease in the number of clients that will be selected at a given round. Unfortunately, this type of strategy is irrelevant if the aim is to distinguish honest clients from malicious ones. In addition, a high number of rounds is needed to achieve a successful selection.

In this section, we develop another sampling strategy based on cluster analysis, which is the technique of grouping clients into clusters such that each client is similar to the clients in the cluster assigned to it, and different from clients in any other cluster. The basic idea is to use the clustering technique to group honest clients in one cluster and malicious nodes in another. Formally, with 2 clusters, given the SV of all clients ${\bf\overline{sv}} = (\overline{sv}_1, \overline{sv}_2,\cdots, \overline{sv}_N)$, find centroids $\mu_1^*$ and $\mu_2^*$ such that
\begin{equation}
\sum_{i\in \mathcal{S}} min_{\mu\in \{\mu_1,\mu_2\}} (\overline{sv}_i - \mu)^2.
\label{cluster}
\end{equation}
Since the SV of clients at the beginning of the training can be highly clustered, 
{we need to be more strict at the beginning of the training, to ensure that malicious clients do not impact too heavily our starting weights.
In our selection strategy, we consider a regularized version of $k$-Means clustering where, instead of providing the number of clusters $k$, we specify a penalty per cluster and minimize the clustering plus the penalty for adding a cluster. In FL, we are only interested in the case where the number of clusters varies between 1 and 2. Formally, the optimization problem (\ref{cluster}) becomes 
 \begin{equation}
    \min\Big( \min_{\mu} \sum_{i\in \mathcal{S}} (\overline{sv}_i - \mu)^2, \sum_{i\in \mathcal{S}} min_{\mu\in \{\mu_1,\mu_2\}} (\overline{sv}_i - \mu)^2+ \lambda\Big),
    \label{cluster2}
\end{equation}
where $\lambda$ $\in [-1,1]$ is the cost of adding a cluster. The value of $\lambda$ is useful for determining how far SV clients must be in order to move from one cluster to two clusters. An increase in $\lambda$ means that the server strategy is more conservative when it comes to detecting malicious nodes. Grouping the clients into two clusters is optimal if the SV of clients satisfies 
\begin{equation}
    \sum_{i\in \mathcal{S}} min_{\mu\in \{\mu_1,\mu_2\}} (\overline{sv}_i - \mu)^2 \leq (1 - \lambda)N \sigma^2 ,
    \label{cluster3sug}
\end{equation}


where $\sigma^2 = \frac{1}{N} \sum_{i\in \mathcal{S}} (\overline{sv}_i - \mu_0)^2$ and $\mu_0 = \frac{1}{N} \sum_{i\in \mathcal{S}} \overline{sv}_i$. The optimal solution of (\ref{cluster}) is obtained as follows : (i) Sort the SV of clients: $\overline{sv}_{\gamma(1)} \leq \overline{sv}_{\gamma(2)}\leq \cdots\leq \overline{sv}_{\gamma(N)}$, where $\gamma(i)$ is the client in $i$-th position in ascending order; (ii) Let $C(i,j) =\sum_{k=i}^j (\overline{sv}_{\gamma(k)} - \mu)^2$ be the cost of grouping $\{\overline{sv}_{\gamma(1)},\dots,\overline{sv}_{\gamma(j)}$ into one cluster with the optimal choice of centroid, $ \mu = \frac{1}{j+1-i} \sum_{k=i}^j \overline{sv}_{\gamma(k)}$, the mean of the points; (iii) Find the last client $\gamma(j^*)$ in ascending order that can be included in the first cluster such that $j^*=\arg\max_{j} \{C(1,j)+C(j+1,N)\}$; (iv) The optimal solution of (\ref{cluster2}) has 2 clusters if 
 \begin{equation}
C(1,j^*)+C(j^*+1,N) \leq N \sigma^2 -\lambda,
\label{cluster4}
\end{equation}
which means that only clients $\{\gamma_{(j^*+1)},..,sv_{\gamma(N)}\}$ are chosen for computing the global model. If inequality (\ref{cluster4}) is not satisfied, all clients are grouped into a single cluster, meaning that no malicious nodes have yet been detected and the global model will use all client models.

\begin{algorithm}[ht!]
    \caption{$ \textsc{ClusFed}$: Client Selection Strategy \label{ClusFed}}
    \begin{small}
        \begin{algorithmic}
            \STATE \textbf{INPUTS:} parameters: $\mathcal{S}$ , ${\bf \overline{sv}}$, $\lambda$ and round $t$.
            \STATE Sort the SV of clients:  $\overline{sv}_{\gamma(1)} \leq \overline{sv}_{\gamma(2)}\leq \cdots\leq \overline{sv}_{\gamma(N)}$.
            \FOR{ $j=1,..,N$}
                \STATE Calculate $C(1,j)$ and $C(j+1,N)$ 
            \ENDFOR
            \STATE Find $j^*$ such that $j^* = \arg\max_{j} \{C(1,j)+C(j+1,N)\}$
            \IF{$C(1,j^*)+C(j^*+1,N) \leq N \sigma^2 -\lambda$}
                \STATE $\mathcal{S}^t = \{\gamma(j^*+1),\cdots, \gamma(N)\}$
            \ELSE 
                \STATE $\mathcal{S}^t = \mathcal{S}$
            \ENDIF
            \STATE \textbf{Return} $\mathcal{S}^t$.
        \end{algorithmic}
    \end{small}
\end{algorithm}

\section{Experimental Evaluation}
\subsection{Dataset and data repartition}
 For all experiments, we use $N=20$ (all participating in each training round) out of which $N_m$ ($4 \leq N_m \leq 10$) nodes are malicious. We evaluate the performance of FedSV based on a digit recognition task called MNIST~\cite{deng2012mnist}. For lack of space, we do not describe the configuration and results of CIFAR-10 in this paper. The {MNIST} dataset consists of 60,000 28x28 pixel grayscale images of handwritten digits from 0 to 9. 10,000 samples are used as a test set, and the remaining 50,000 are used for training. 
The dataset contains 10 different classes. For MNIST, we use a CNN containing 2 convolution layers with respectively 32 and 64 filters, each followed by a ReLU correction. We then have a max pooling layer, followed by a 0.25 probability dropout, a linear layer with ReLU, a 0.5 probability dropout followed by a final linear layer with softmax. For the optimizer we had the best results using AdaDelta, however, it has the disadvantage of hindering late detection of malicious clients because the gradients are less significantly updated when the number of rounds increases. 

We divide the clients into $m$ groups with $m$ being the number of classes in our dataset. Client $i$ is part of group $L_k$ with $k = i\%m$. Each client of group $L_k$ receives samples from the class $k$, $(k+1)\%m$ and $(k+2)\%m$. Therefore, each client has data points from only 3 classes. We decided to focus on non-IID  cases because they are the most challenging and realistic scenarios. Hence, each client has only a portion of the data and performs worse than in an IID case, which also makes learning more difficult. Indeed, clients that don't perform well due to the distribution of data can be confused with malicious clients, since they also don't provide us with enough information to distinguish malicious clients. This is why, when datasets of clients are  extremely heterogeneous, even insignificant attacks can become very effective. 

\subsection{Attacks}
In our evaluation, we explore the impact of three main attacks, namely Gaussian noise, Backdoor, and Sign-Flipping attacks. Due to space constraints, we will only present the attack based on the Sign-Flipping attack, which was the most impactful attack out of the three, but the results are qualitatively similar for the two other attacks.  The sign flipping attack consists of switching the sign of the weights. By doing so, one alters the performance of the model without changing the magnitude of the weights, making the attack difficult to detect for some defenses~\cite{li2020learning}.

\subsection{Defenses}
\label{sec:def}
We compare FedSV with three other common defense strategies, namely Multi-Krum~\cite{Blanchard2017Machine}, Trimmed Mean~\cite{Pillutla2022Robust} and Median~\cite{Pillutla2022Robust}. 
Multi-Krum defense is further split in two, Multi-KrumF and Multi-Krum. Multi-KrumF corresponds to the defense in a full knowledge setting, where  the number of malicious clients is known. Multi-Krum is the name of the defense in a context of partial knowledge, where the exact number of malicious nodes is unknown.We also evaluate the no defense federated averaging (FedAvg)~\cite{McMahan2016CommunicationEfficientLO} with malicious nodes and without the malicious nodes as a baseline. This allows us to see, in terms of accuracy, how close FedSV's performance is to the solution obtained by FedAvg without malicious nodes. Note that the FedAvg does not consider the security support in the learning process.


\subsection{Experimental Results}
We run 20 simulations for each series of 100 rounds and for all algorithms.  We set the learning rate to $\eta=0.005$, $\beta = 0.7$, $\lambda = 0$ and $E=5$. In partial knowledge setting, we test Multi-Krum with the hypothesis that 50\% of the nodes are malicious.
 
\begin{figure}[htb!]
	\includegraphics[width=\linewidth]{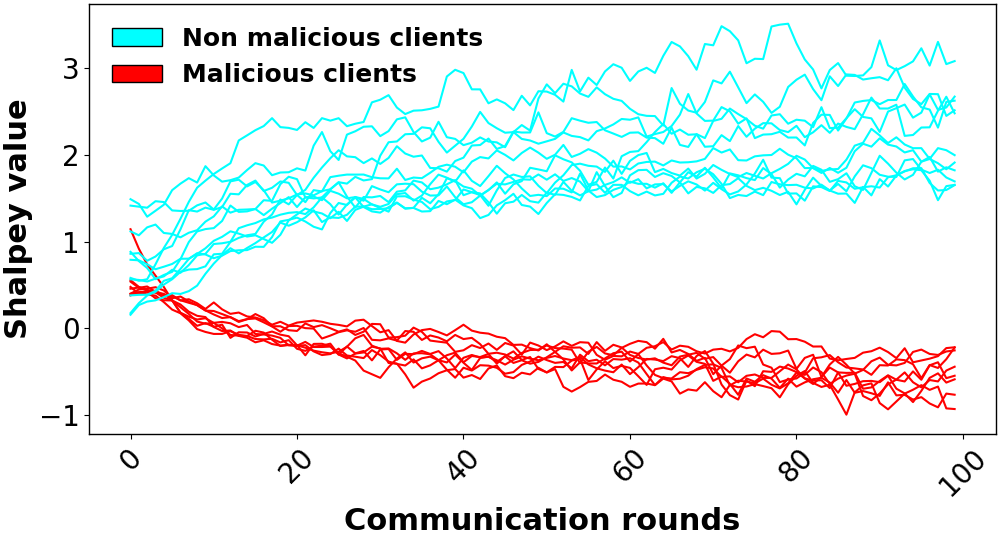}
	\caption{SV of all clients  during the training rounds.}
	\label{fig:Sv-evol}
\end{figure}

In Fig.~\ref{fig:Sv-evol}, we represent the SV of all clients, specifying that malicious clients are marked in red and  healthy clients in cyan.  We observe that the SV  is a good measure for distinguishing malicious from non-malicious clients, even in the early stages of training. 
Therefore, using SV, FedSV is able to detect malicious clients and exclude them from participation in the global model unless if they show better results thanks to their SV.  We note that the SV can be calculated at each round to give clients  a chance to return to training, or at any other frequency to balance runtime and defensive strength.

\begin{figure}[htb!]
	\includegraphics[width=\linewidth]{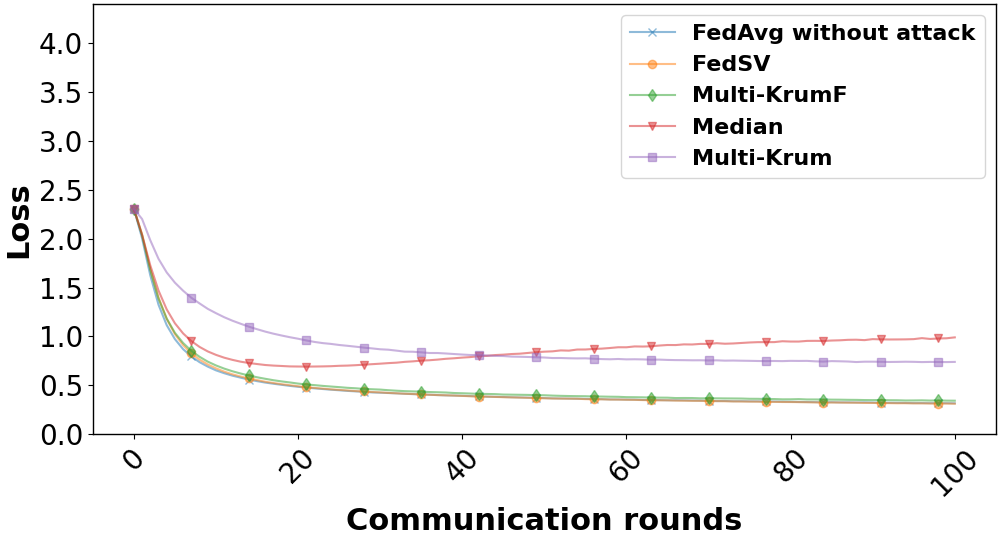}
	\caption{Global loss  in the presence of 40\% malicious nodes.}
	\label{fig:loss40}
\end{figure}

Fig.~\ref{fig:loss40} and~\ref{fig:loss55} shows average loss during 100 communication rounds for the five most performant strategies among the 8 we tested. In the presence of 40\% malicious nodes FedSV and Multi-KrumF have a similar loss to the baseline strategy FedAvg without attack. Multi-Krum has a notably higher loss but still converges, Median has the worse loss and does not converge in 100 rounds.
In presence of 50\% of malicious nodes both Multi-Krum based defense do not achieve to improve their global model during the communication rounds. For Median's loss we have a similar behavior as for 40\% it first decreases before increasing again due to the perturbation of the malicious nodes. FedSV still shows satisfying convergence properties having a similar behavior than the baseline. Under 50\% of malicious clients only FedSV remains a robust defense.

\begin{figure}[htb!]
	\includegraphics[width=\linewidth]{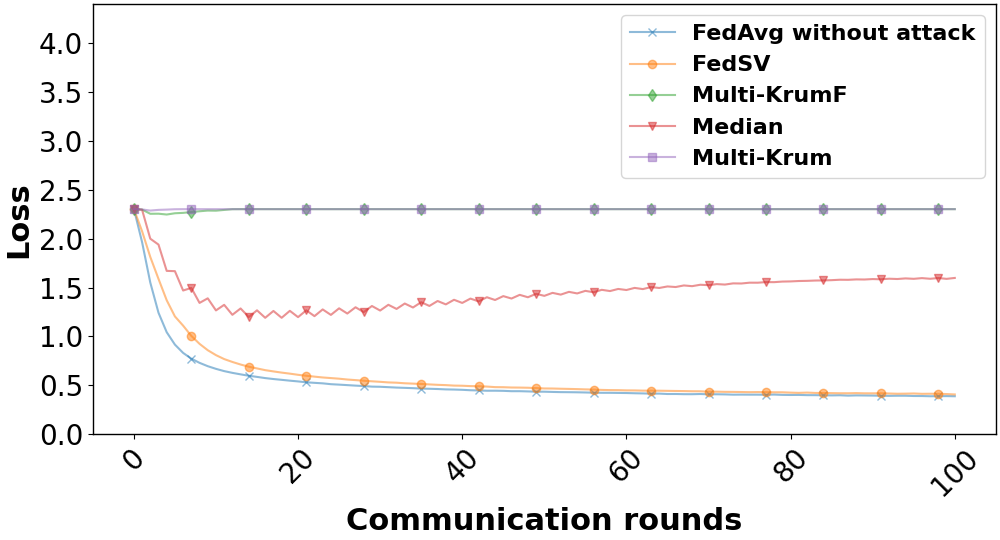}
	\caption{Global loss  in the presence of 55\% malicious nodes.}
	\label{fig:loss55}
\end{figure}



Fig.~\ref{fig:acc} shows the system's accuracy for each strategy defined in Sec.~\ref{sec:def}. Comparing the results to the case without attack, we see that FedSV has similar accuracy results, even when increasing  the proportion of malicious clients up to 50\%. Many defenses struggle to deliver good results as the number of malicious clients increases, and often become useless when more than half the clients are malicious. This is not the case with FedSV, as it calculates the SV of each client based on the performance of its models when tested against a set of test data.  We observe that FedSV is robust to attacks from malicious nodes even if their proportion is more than half. 

\begin{figure}[htb!]
	\centering
	\includegraphics[width=0.8\linewidth]{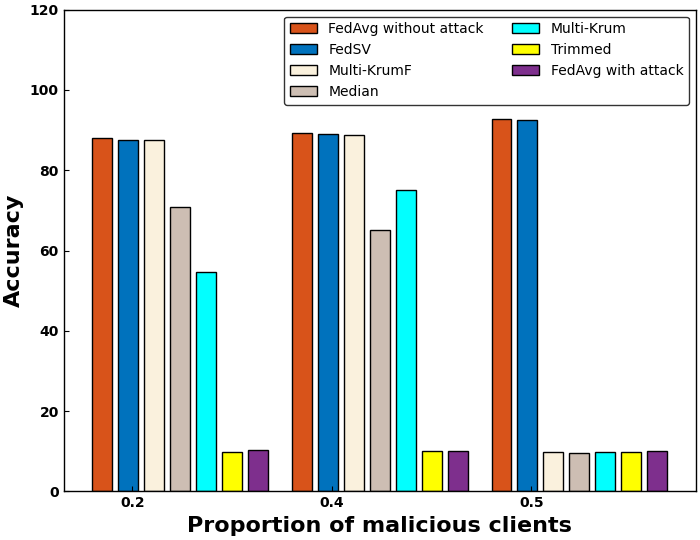}
	\caption{Accuracy  comparison on different proportions of malicious clients.}
	\label{fig:acc}
\end{figure}

In a setup where we precisely know the number of malicious clients Multi-KrumF also has a high accuracy, but we observe that the strategy has a breaking point around 50\% of malicious nodes, with healthy nodes being a minor the strategy is not able anymore to distinguish between the malicious and non-malicious nodes. We note that the assumption about the knowledge of malicious nodes needed by Multi-KrumF is not realistic, especially for federated learning. 

\begin{figure}[htb!]
	\centering
	\includegraphics[width=0.8\linewidth]{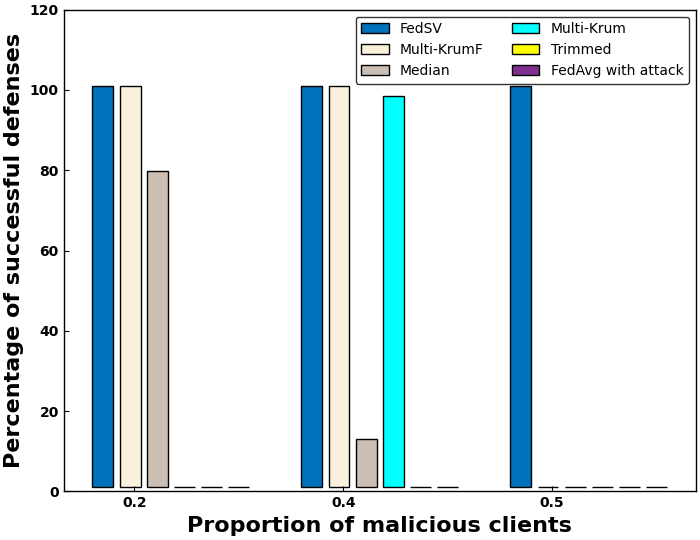}
	\caption{Percentage of successful defences under each strategy.}
	\label{fig:success-attacks}
\end{figure}

We observe that in a more realistic setup, Multi-Krum performs significantly worse than in the previous one. More surprisingly, when the proportion of malicious clients is 0.2, the strategy obtains a lower accuracy than with a proportion of 0.4. Due to the heterogeneity of the data distribution and to the assignment mechanism for malicious clients, we have half of the instances from classes 0, 1, 2 that are held by the malicious nodes. Therefore, the remaining nodes with data samples from classes 0, 1, and 2 are never selected by the system because their weights are far from the weights of the other clients. Thus, the system does not select any samples for those three classes. Leading to a lower accuracy and higher loss.


In Fig.~\ref{fig:success-attacks}, we examine  the accuracy obtained for each run. Here, we assume  a successful defense must have at least 80\% of the accuracy obtained with FedAvg without an attacking baseline. We observe  that FedSV remains very robust and its accuracy is not affected in any run. In the other solutions, the attack is able to achieve a very high success rate, reducing accuracy by at least 20\% compared with the accuracy obtained by FedAvg without attack.






\section{Conclusion}  In this paper, we proposed a new framework for distinguishing malicious from healthy clients  based on Shapley value, and showed that FedSV is able to identify malicious clients early in the learning process. Furthermore, based on this framework we proposed a selection strategy called ClusFed that automatically and successfully prunes the malicious nodes during training, leading to very similar performance results as the baseline.  

Two interesting directions for future work are the design of new SV approximation frameworks to handle a larger number of clients in a reasonable time and the design of new selection strategies that can, in addition to pruning malicious nodes, find the optimal combination of clients at each round (e.g., not necessarily select all non-malicious clients) and thus perform better than the baseline.

\bibliographystyle{IEEEtran}
\bibliography{bibliography} 
\end{document}